# Enhancing Acute Kidney Injury Prediction through Integration of Drug Features in Intensive Care Units

Gabriel D. M. Manalu[1], Mulomba Mukendi Christian[2*], Songhee You[3], Hyebong Choi[4]

[1]*Undergraduate student, Department of Computer Science and Electrical Engineering, Handong Global University, Korea*
*22000240@handong.ac.kr*
[2*]*Doctoral student, Department of Advanced Convergence, Handong Global University, Korea*
*mulombachristian@handong.ac.kr*
[3]*Assistant professor, School of Spatial Environment System Engineering, Handong Global University, Korea*
*shyou@handong.edu*
[4]*Associate professor, School of Global Entrepreneurship and ICT, Handong Global University, Korea*
*hbchoi@handong.edu*

## *Abstract*

*The relationship between acute kidney injury (AKI) prediction and nephrotoxic drugs, or drugs that adversely affect kidney function, is one that has yet to be explored in the critical care setting. One contributing factor to this gap in research is the limited investigation of drug modalities in the intensive care unit (ICU) context, due to the challenges of processing prescription data into the corresponding drug representations and a lack in the comprehensive understanding of these drug representations. This study addresses this gap by proposing a novel approach that leverages patient prescription data as a modality to improve existing models for AKI prediction. We base our research on Electronic Health Record (EHR) data, extracting the relevant patient prescription information and converting it into the selected drug representation for our research, the extended-connectivity fingerprint (ECFP). Furthermore, we adopt a unique multimodal approach, developing machine learning models and 1D Convolutional Neural Networks (CNN) applied to clinical drug representations, establishing a procedure which has not been used by any previous studies predicting AKI. The findings showcase a notable improvement in AKI prediction through the integration of drug embeddings and other patient cohort features. By using drug features represented as ECFP molecular fingerprints along with common cohort features such as demographics and lab test values, we achieved a considerable improvement in model performance for the AKI prediction task over the baseline model which does not include the drug representations as features, indicating that our distinct approach enhances existing baseline techniques and highlights the relevance of drug data in predicting AKI in the ICU setting.*

Corresponding Author: mulombachristian@handong.ac.kr
Tel: +82 10-3506-2882, Fax: +82 54-260-3409
Doctoral student, Department of Advanced Convergence, Handong Global University, Korea

## 1. Introduction

Acute kidney injury (AKI) is one of the major complications that occur in the intensive care unit (ICU) setting, often incurring additional costs such as extended length of stay [1-3] and increased mortality rate from consequent chronic kidney disease (CKD) or cardiovascular diseases [4, 5]. Based on the Kidney Disease Improving Global Guidelines (KDIGO) definition, AKI is determined by an increase in serum creatinine or a decrease in urine outpour and can be split into several stages depending on the severity [6]. Despite the significance of identifying AKI, the main obstacle for nephrologists and ICU clinicians is the absence of early prediction and effective disease management [7]. Therefore, there is a pressing need for early risk prediction and timely therapeutic interventions for AKI in the ICU setting.

In recent years, there has been a growing trend to analyze ICU data to either try to derive meaningful insights such as in the case of a comorbidity analysis [8], or to predict specific outcomes which occur in the ICU. One of the most common prediction tasks is to investigate the incidence of AKI based on data found in an ICU's electronic health records (EHR) [9, 10] due to its detailed nature, containing data such as patient demographics, vital signs, lab values, and comorbidities in a specific ICU stay. Medical Information Mart for Intensive Care (MIMIC-III) version 1.4 is one such large EHR dataset which contains various kinds of data related to a patient's ICU stays, such as demographic and lab test result data among others, including urine output and serum creatinine values which are standard markers for AKI. Numerous studies investigating AKI prediction in the MIMIC-III dataset is based on using machine learning and deep learning methods [11-21], aiming to either develop accurate predictive models for AKI or explore existing modalities or features in the dataset to improve the accuracy of state-of-the-art models.

One such modality that has the potential to improve current models to predict AKI is the utilization of a patient's prescription data throughout their ICU stay. Nephrotoxicity, which refers to any renal injury resulting from the direct or indirect effects of medication, ranks as the third leading cause of AKI [21]**.** Several papers have investigated the interplay between the administration of nephrotoxic drugs and the heightened risk of developing AKI in the ICU [23]. Research indicates a plausible causal relationship, suggesting that certain medications, when administered in specific doses or over prolonged durations, can substantially compromise renal function, leading to the onset or exacerbation of AKI. Therefore, gaining a comprehensive understanding of the impact of these drugs on renal health is crucial for devising effective strategies to mitigate the risk of AKI in clinical practice. In line with this, the extensive MIMIC-III dataset also contains abundant prescription data for each patient (including the drug name, generic drug name, National Drug Code (NDC), as well as start and end times of the prescription order among others). Incorporating diverse drug representations of those prescribed medications, such as molecular fingerprints, into our dataset can possibly enhance the predictive performance of AKI, similar to their successful application in other studies focusing on drug property prediction [24, 25]**.**

In light of the previous research conducted in this area, we propose a study which uses machine learning classifiers applied to predict AKI incidence in a dataset with baseline cohort features and another dataset with the same baseline cohort features as well as the addition of drug embeddings. The findings from the model demonstrate the enhancement of AKI prediction through the incorporation of drug embeddings and other patient cohort features. Furthermore, we examine the application of drug representations, specifically extended-connectivity fingerprint (ECFP) [26]**.** In the end, we believe that this research will give key contributions in three different aspects. First, the study examines the process of extracting relevant data from an EHR database with the purpose of AKI prediction, as well as creating drug features from the relevant extracted data. In this process, we try to emphasize the importance of drug features as a modality to consider in the prediction of other studies' clinical tasks. Second, our suggested approach of including drug representations show improvement in the task of AKI prediction when used on top of the standard patient cohort features. Even though many other works have tried to use machine learning and deep learning methods to try to predict AKI, to the best of our knowledge this research is among the first in the healthcare domain to employ drug embeddings, focusing on the chemical structure of drugs, for this specific task. By conducting this study, we believe that our work can be the building block for further research to address the causes of drug-induced AKI.

### 1.1 Related Work

There have been numerous studies which tries to predict AKI in large-scale EHR datasets similar but not limited to MIMIC-III. Among them, as reviewed by Yu *et al.* [27], both machine learning and deep learning methods were used. One of the earlier studies, conducted by Zimmerman *et al.* [13], investigated the application of machine learning and deep learning methods, in particular logistic regression, random forest, and neural networks, while using only the MIMIC-III patients' demographic, vital signs, and lab data. As one of the studies pioneering the use of machine learning and deep learning on features extracted from an ICU stay, their work showed that simply using demographic and physiologic features can come up with competitive results in AKI prediction. Another study by Liang *et al.* collected patients' data from three databases (MIMIC-III, AmsterdamUMCdb [28] and their local SHJZU-ICU database) to generalize the AKI prediction problem across databases of different locations comprising different ethnicities, strictly using machine learning methods. Their study made a great contribution in the field of AKI research using machine learning, as they not only developed machine learning models from a transcontinental cohort, but they also validated their models on a local database, returning comparably good outcomes.

Out of the existing research which use data collected in the ICU to predict AKI, some have explored a multimodal approach as additional features to be used for prediction. The specific data modality that numerous studies have elected to use on top of standard patient cohort feature are a patient's clinical notes, since databases such as MIMIC-III contain a patient's record of clinical notes, Sun *et al.* used natural language processing on clinical notes as well as physiological measurements to predict AKI developed within 72 hours by utilizing machine learning algorithms [16]. Through their work, they have shown that the addition of another modality of data, specifically unstructured features such as clinical notes, can improve the risk identification of AKI onset when effectively applied. However, in our study, rather than incorporating clinical notes as an extra element, drug embeddings are employed within a multimodal approach.

## 2. Experiments

### 2.1 Data

The data utilized in this research originates from the aforementioned open dataset MIMIC-III [29]. Out of the 46,520 patients and 61,532 ICU stays in this database, we want to filter the ICU stay cohort into only those which does not meet the exclusion criteria, as well as those where a drug has been consumed within the first 24 hours following ICU admission. To start, patients who developed AKI 72 hours after admission were excluded. Additionally, patients who had a history or was diagnosed with AKI or chronic kidney disease (CKD) before admission were excluded, by checking their related comorbidities. Then, the admission age for each patient in the cohort was calculated and those who were below 18 years of age were excluded. After applying these exclusion criteria, the patient's demographics, maximum, mean, and minimum values of first-day vitals, maximum and minimum values of first-day lab test values, first-day urine output and a binary value to indicate whether a patient was mechanically ventilated or not were derived as the features for the ICU stay cohort, which had 23,581 patients and 29,623 ICU stays. In total, there were 72 extracted features and we will then refer to those extracted features as "cohort features".

We further screen the ICU stay cohort based on whether or not a drug was prescribed within the first 24 hours (first day) after a patient's admission, which is a similar time frame to the features which we extracted beforehand. We identified whether a patient was prescribed a drug within the first 24 hours by checking the start date of the drug prescription, ensuring that it is a date after but within 24 hours of the patient's admission. After screening, we end up with a cohort of 16,513 patients and 19,073 ICU stays.

**Table 1. Summary statistics of the number of patients and ICU stays used for the study**

| | *# of Patient* | *# of ICU stays* |
|---|---|---|
| MIMIC-III | 46,520 | 61,532 |
| MIMIC-III (after applying exclusion criteria) | 23,581 | 29,623 |
| Final Cohort (after checking drug prescription) | 16,513 | 19,073 |

### 2.1.1 Handling Missing Values.

After screening the ICU stay cohort based on whether or not a drug was prescribed within the first 24 hours after admission, the statistics for the datasets after each step of modification are shown in Table 1. Next, we are left with the final cohort containing the relevant features that we will use for the classification tasks. In order for machine learning classifiers to use the structured clinical variables that have been obtained, the variables with missing values can be removed or those missing values can be imputed. Out of the 72 cohort features in the final cohort, we removed those features which had over 20% missing values. After such variables were removed, we remain with 63 variables on which we should perform imputation to handle the missing values. For our imputation method, we have chosen Multivariate Imputation with Chain-Equation, or MICE imputation [30] to fill in those missing variables. MICE imputation makes the assumption that data is missing-at-random, and creates a regression model to impute the values of a variable while using other variables as the predictors. In a sense, MICE handles missing data by recognizing that the features can be interdependent and tries to use those interdependencies to make accurate imputations.

### 2.1.2 Drug to SMILES Pipeline.

Some subsequent that needs to be undertaken after imputing the missing values are extracting the prescription information and converting each drug to its corresponding drug embedding, or molecular fingerprint in this case. Our work tries to follow a pipeline which was used in a similar paper that converts prescription information ultimately to ECFP and pre-trained SMILES Transformer [31]. The first step in the pipeline is that for each of the ICU stays where a drug was prescribed within the first 24 hours after admission, prescription information is extracted, which specifically consists of the drug name, the generic drug name, and the National Drug Code (NDC). Altogether, there are 1,794 unique drugs, 2,871 unique NDC codes, and 1,416 unique generic drug names in the cohort. Then, the next step is obtaining the Simplified Molecular Input Line Entry System (SMILES) notation [32], a text-based representation of the drug's complex chemical structures. Since the molecular fingerprint that we will use as drug representations in this use SMILES representation as the basis, those drugs which do not have a SMILES representation will be removed from the final dataset used for prediction.

We first try to find a drug's SMILES representation by searching the drug's generic name in the Pubchem [33] compound database utilizing the Python pubchempy library and querying if there is a SMILES representation associated with it. However, this may not yield results since there are a lot of generic drug names which are not standardized making it impossible for them to be recognized in the Pubchem database. If this occurs, we try to find the drug's SMILES representation using its drug name. As a last measure, we use the drug's NDC code and find its corresponding proprietary name in the Food and Drug Administration (FDA) database. This corresponding proprietary name will then be queried in the Pubchem database in the same way the drug generic name and drug name was queried to obtain the SMILES representation. In existing literature, various methods leverage discrete molecular representation, such as SMILES strings, and transform them into lower-dimensional continuous representations. Our study tries to use the ECFP vector representation. This drug representation is integrated with other aforementioned cohort features to enhance the effectiveness of the machine learning model's prediction by adding another modality to the data.

### 2.1.3 Representations of Drugs in Latent Space.

In the field of cheminformatics, encoding molecular descriptors and fingerprints into low-dimensional vectors is a common practice. Developing an effective representation for drugs in the molecular space is a complex task, given the expansive search space and the intricate, unstructured nature of drugs. Various methods have been proposed, such as applying convolutional neural networks on graphs [34] to convert molecular representations into efficient continuous embedding vectors. For this study, we opted to use the widely used SMILES representation for clinical drugs, transforming it into a distinct vector representation. Initially, we utilized the RDKit 3 open-source cheminformatics library to convert the SMILES string into ECFP, a prevalent type of molecular fingerprint. ECFP generates a fixed-length (1024 in this case) binary vector, where the presence of the designated substructure is indicated by 1, and its absence by 0, employing a predetermined hash function on the concatenated neighborhood features.

## 2.2 Machine Learning Classifiers

Ensemble learning algorithms, specifically different boosting and bagging algorithms, were used as machine learning classifiers for predicting AKI incidence. Out of the bagging algorithms, Random Forest [35] was chosen because of its high predictive capability and interpretability despite its simplicity to use. Meanwhile, XGBoost [36] excels as one of the preferred boosting algorithms for maximizing predictive accuracy. Another algorithm that is efficient and is known to work well for high-dimensional data, LightGBM [37] stands out for its computational efficiency and effective handling of large datasets. CatBoost [38] which stands for categorical boosting, is another boosting algorithm which is notable for its handling of categorical variables, removing the need for extensive preprocessing while maintaining efficiency. For each machine learning classifier, a specific experimental setup is defined and the best results are shown in further parts of the paper.

## 2.3 Experimental Setup

To evaluate how well the machine learning classifiers can predict AKI, three metrics are recorded, which are: Area Under the Receiver Operating Characteristics (AUROC), Area Under Precision-Recall (AUPRC), and F1 score. AUROC measures the model's ability to distinguish between the positive and negative class across different probability thresholds. AUPRC evaluates the trade-off between precision, or the positive predictive value, and recall, or the capability of a model to capture the positive instances. Lastly, F1 score is the harmonic mean of precision and recall and gives a great insight for a model's performance, especially if there is an imbalance between the positive and negative classes. In terms of the hyperparameters to tune for each machine learning classifier, multiple empirical experiments were run and the appropriate hyperparameters to get the best results were selected.

## 3. Results

Since the purpose of this study is to demonstrate the impact of integrating drug representations for the AKI prediction task, the machine learning classifiers were first trained only on the dataset which contains the baseline cohort features, without the addition of ECFP fingerprints. All machine learning classifiers had the same hyperparameter values when processing the cohorts before and after addition of the drug embeddings, and the results of the proposed multimodal approach are measured against the results of the model on the baseline cohort features. Out of all the machine learning classifiers with empirically set hyperparameters, CatBoost outperformed and got considerably higher improvement of AUROC, AUPRC, and F1 scores when applied to the AKI prediction task on the cohort with the ECFP fingerprint as additional features. Though the other machine learning classifiers had the potential to increase the AUROC, AUPRC, and F1 scores after the addition of drug embeddings, the observed improvement was minute and therefore was not reported due to the poor results. This difference between the perceived improvement of scores after integrating drug embedding between CatBoost and other ensemble learning algorithms can be due its different inherent and its treatment of features which suits better for the drug embeddings.

Based on the findings in Table 2, there is an increase of around %2.5 AUROC, %4.5 AUPRC, and %4.9 F1 score. The noticeable contrast in model performance distinctly indicates that the advocated multimodal approach exhibits superior predictive capability. Also, this enhancement in all metrics after the addition of drug embeddings suggest that those embeddings can be key to making to the prediction of AKI incidence for a patient in an ICU stay.

**Table 2. CatBoost results for the task of predicting AKI incidence**

| *Features* | *Drug Embedding* | *AUROC* | *AUPRC* | *F1 Score* |
|---|---|---|---|---|
| Baseline Cohort Features | - | 0.679 | 0.420 | 0.298 |
| Multimodal Approach | ECFP | 0.704 | 0.465 | 0.348 |

## 4. Discussion

The model outcomes outlined in Table 2 shows the models' outcomes, which underscore the advantages of incorporating drug embeddings as an additional feature. Through our experiments, we validated our hypothesis that drug embeddings can improve AKI prediction using data taken from the ICU. The proposed multimodal techniques, integrating drug embeddings with baseline cohort features consistently outperform all models which do not use those drug embeddings, demonstrating superior performance. To our knowledge, this study is the first to leverage molecular representations of clinical drugs for predicting AKI incidence, especially using MIMIC-III data. This multimodal approach represents a novel contribution the field of AKI research, introducing the usage of a new modality that has been overlooked.

The experimental results provide compelling evidence that support the methodology of using the multimodal approach with drug embeddings as one of the modalities for predicting AKI. As shown in Table 2, the utilization of ECFP fingerprints as features have shown promise, increasing the scores in the selected metrics. However, for future experiments there can be different types of drug embeddings, or ways to represent the drugs consumed by a patient in the molecular level, and they can be evaluated in a similar fashion, being compared to each other. We contend that leveraging drug information in this manner may open avenues for novel approaches in AKI prediction.

Additionally, our literature review showcased works which tried to predict AKI using MIMIC-III data with baseline features and other modalities. Zimmerman *et al.* [13] tried to predict AKI using features similar to the baseline cohort features which we have used in this study, which are patients' demographic, vital signs, and lab test results. Other studies reviewed have elected for a multimodal approach, in their cases using clinical notes as an added modality. Sun *et al.* [16] used natural language processing on clinical notes as another modality in addition to the baseline cohort features, noting the improvement after integration of the clinical notes. In comparison, we formulated some exclusion criteria to extract our own cohort from the MIMIC-III database and have additionally extracted prescription data to create the drug embeddings for improved AKI prediction. This targeted approach, coupled with the extraction of prescription data to create drug embeddings, has not only refined our cohort but has also introduced a novel modality, distinct from previous multimodal studies. This differentiation sets our work apart, offering another perspective on AKI prediction with the integration of drug-specific information.

## 5. Conclusion

In conclusion, our paper not only addresses the pressing need for early risk prediction and therapeutic intervention in AKI but also introduces a groundbreaking multimodal approach that leverages drug embeddings for enhanced predictive accuracy. This work not only fills a critical gap in the current understanding of AKI prediction but also sets the foundation for future investigations into the intricate relationship between drug-induced nephrotoxicity and AKI onset in critical care settings. We used the openly available MIMIC-III for the EHR dataset, and based on existing literature, we generated a distinct ICU stay cohort based on multiple exclusion criteria. To complete the multimodal approach, we extracted prescription information and processed them through a pipeline in order to ultimately convert them to drug embeddings, specifically ECFP molecular fingerprints. After several experiments using machine learning classifiers, the data and results show that CatBoost returns the best results in this multimodal approach for the task of AKI incidence prediction.

Even though our hypothesis has been proven by the increased performance of the multimodal approach over the baseline approach, our research has given light to several limitations we need to consider. First, it was difficult to process every drug that we needed to extract from the MIMIC-III prescription information, due to the informal nature of the data. For example, some of the names of the drugs might be misspelled and manual conversion of the drug's name and generic name might be needed, even after using an intricate drug preprocessing pipeline. Second, it is hard to interpret the model results, especially because of the complex nature of molecular fingerprints.

The future work in this research area, then, might follow up on these directions. First, the drug processing pipeline can be improved after an extensive investigation by manually changing more drug names to get more compounds with their molecular fingerprints. Second, the featurizers which are used to create the ECFP fingerprints can be analyzed and each substructure or column in the ECFP fingerprint can be explored more deeply, especially if they are more prevalent among positive or negative AKI ICU stays.

## Acknowledgement

This work was supported by the MSIT(Ministry of Science and ICT), Korea, under the National Program for Excellence in SW) supervised by the IITP(Institute of Information & Communications Technology Planning & Evaluation) in 2023(2023-0-00055).

## References

[1] C. Hobson *et al.*, “Cost and Mortality Associated With Postoperative Acute Kidney Injury,” *Ann. Surg.*, vol. 261, no. 6, pp. 1207–1214, Jun. 2015, doi: 10.1097/SLA.0000000000000732.
[2] S. A. Silver, J. Long, Y. Zheng, and G. M. Chertow, “Cost of Acute Kidney Injury in Hospitalized Patients,” *J. Hosp. Med.*, vol. 12, no. 2, pp. 70–76, Feb. 2017, doi: 10.12788/jhm.2683.
[3] S. A. Silver and G. M. Chertow, “The Economic Consequences of Acute Kidney Injury,” *Nephron*, vol. 137, no. 4, pp. 297–301, 2017, doi: 10.1159/000475607.
[4] S. G. Coca, B. Yusuf, M. G. Shlipak, A. X. Garg, and C. R. Parikh, “Long-term Risk of Mortality and Other Adverse Outcomes After Acute Kidney Injury: A Systematic Review and Meta-analysis,” *Am. J. Kidney Dis.*, vol. 53, no. 6, pp. 961–973, Jun. 2009, doi: 10.1053/j.ajkd.2008.11.034.
[5] R. Wald, “Chronic Dialysis and Death Among Survivors of Acute Kidney Injury Requiring Dialysis,” *JAMA*, vol. 302, no. 11, p. 1179, Sep. 2009, doi: 10.1001/jama.2009.1322.
[6] “Section 2: AKI Definition,” *Kidney Int. Suppl.*, vol. 2, no. 1, pp. 19–36, Mar. 2012, doi: 10.1038/kisup.2011.32.
[7] Y. Wang, Y. Fang, J. Teng, and X. Ding, “Acute Kidney Injury Epidemiology: From Recognition to Intervention,” in *Contributions to Nephrology*, vol. 187, X. Ding and C. Ronco, Eds., S. Karger AG, 2016, pp. 1–8. doi: 10.1159/000443008.
[8] H. Sookyung, N. Cheryl, "Comorbidity Analysis on Big Data", *International Journal of Advanced Culture Technology*, vol. 7 no. 2, pp. 13-18, Jun. 2019, doi: 10.17703/IJACT.2019.7.2.13.
[9] J. Case, S. Khan, R. Khalid, and A. Khan, “Epidemiology of Acute Kidney Injury in the Intensive Care Unit,” *Crit. Care Res. Pract.*, vol. 2013, pp. 1–9, 2013, doi: 10.1155/2013/479730.
[10] P. Pickkers *et al.*, “Acute kidney injury in the critically ill: an updated review on pathophysiology and management,” *Intensive Care Med.*, vol. 47, no. 8, pp. 835–850, Aug. 2021, doi: 10.1007/s00134-021-06454-7.
[11] K. Shawwa, E. Ghosh, S. Lanius, E. Schwager, L. Eshelman, and K. B. Kashani, “Predicting acute kidney injury in critically ill patients using comorbid conditions utilizing machine learning,” *Clin. Kidney J.*, vol. 14, no. 5, pp. 1428–1435, Apr. 2021, doi: 10.1093/ckj/sfaa145.
[12] Y. Li, L. Yao, C. Mao, A. Srivastava, X. Jiang, and Y. Luo, “Early Prediction of Acute Kidney Injury in Critical Care Setting Using Clinical Notes,” in *2018 IEEE International Conference on Bioinformatics and Biomedicine (BIBM)*, Madrid, Spain: IEEE, Dec. 2018, pp. 683–686. doi: 10.1109/BIBM.2018.8621574.
[13] L. P. Zimmerman *et al.*, “Early prediction of acute kidney injury following ICU admission using a multivariate panel of physiological measurements,” *BMC Med. Inform. Decis. Mak.*, vol. 19, no. S1, p. 16, Jan. 2019, doi: 10.1186/s12911-019-0733-z.
[14] X. Zhang, S. Chen, K. Lai, Z. Chen, J. Wan, and Y. Xu, “Machine learning for the prediction of acute kidney injury in critical care patients with acute cerebrovascular disease,” *Ren. Fail.*, vol. 44, no. 1, pp. 43–53, Dec. 2022, doi: 10.1080/0886022X.2022.2036619.
[15] Q. Liang, Y. Xu, Y. Zhou, X. Chen, J. Chen, and M. Huang, “Severe acute kidney injury predicting model based on transcontinental databases: a single-centre prospective study,” *BMJ Open*, vol. 12, no. 3, p. e054092, Mar. 2022, doi: 10.1136/bmjopen-2021-054092.
[16] M. Sun *et al.*, “Early Prediction of Acute Kidney Injury in Critical Care Setting Using Clinical Notes and Structured Multivariate Physiological Measurements,” *Stud. Health Technol. Inform.*, vol. 264, pp. 368–372, Aug. 2019, doi: 10.3233/SHTI190245.
[17] F. Alfieri *et al.*, “A deep-learning model to continuously predict severe acute kidney injury based on urine output changes in critically ill patients,” *J. Nephrol.*, vol. 34, no. 6, pp. 1875–1886, Dec. 2021, doi: 10.

1007/s40620-021-01046-6.
[18] C. Wei, L. Zhang, Y. Feng, A. Ma, and Y. Kang, "Machine learning model for predicting acute kidney i njury progression in critically ill patients," *BMC Med. Inform. Decis. Mak.*, vol. 22, no. 1, p. 17, Dec. 2 022, doi: 10.1186/s12911-021-01740-2.
[19] N. Sato, E. Uchino, R. Kojima, S. Hiragi, M. Yanagita, and Y. Okuno, "Prediction and visualization of a cute kidney injury in intensive care unit using one-dimensional convolutional neural networks based on ro utinely collected data," *Comput. Methods Programs Biomed.*, vol. 206, p. 106129, Jul. 2021, doi: 10.1016/ j.cmpb.2021.106129.
[20] S. Le *et al.*, "Convolutional Neural Network Model for Intensive Care Unit Acute Kidney Injury Predictio n," *Kidney Int. Rep.*, vol. 6, no. 5, pp. 1289–1298, May 2021, doi: 10.1016/j.ekir.2021.02.031.
[21] N. Tomašev *et al.*, "A clinically applicable approach to continuous prediction of future acute kidney injur y," *Nature*, vol. 572, no. 7767, pp. 116–119, Aug. 2019, doi: 10.1038/s41586-019-1390-1.
[22] G. T. M. Sales and R. D. Foresto, "Drug-induced nephrotoxicity," *Rev. Assoc. Médica Bras.*, vol. 66, no. suppl 1, pp. s82–s90, 2020, doi: 10.1590/1806-9282.66.s1.82.
[23] M. A. Perazella, "Drug use and nephrotoxicity in the intensive care unit," *Kidney Int.*, vol. 81, no. 12, p p. 1172–1178, Jun. 2012, doi: 10.1038/ki.2010.475.
[24] H. Öztürk, A. Özgür, and E. Ozkirimli, "DeepDTA: deep drug–target binding affinity prediction," *Bioinfor matics*, vol. 34, no. 17, pp. i821–i829, Sep. 2018, doi: 10.1093/bioinformatics/bty593.
[25] K. Yang *et al.*, "Analyzing Learned Molecular Representations for Property Prediction," *J. Chem. Inf. Mo del.*, vol. 59, no. 8, pp. 3370–3388, Aug. 2019, doi: 10.1021/acs.jcim.9b00237.
[26] D. Rogers and M. Hahn, "Extended-Connectivity Fingerprints," *J. Chem. Inf. Model.*, vol. 50, no. 5, pp. 742–754, May 2010, doi: 10.1021/ci100050t.
[27] X. Yu, Y. Ji, M. Huang, and Z. Feng, "Machine learning for acute kidney injury: Changing the traditiona l disease prediction mode," *Front. Med.*, vol. 10, p. 1050255, Feb. 2023, doi: 10.3389/fmed.2023.1050255.
[28] P. J. Thoral *et al.*, "Sharing ICU Patient Data Responsibly Under the Society of Critical Care Medicine/E uropean Society of Intensive Care Medicine Joint Data Science Collaboration: The Amsterdam University Medical Centers Database (AmsterdamUMCdb) Example*," *Crit. Care Med.*, vol. 49, no. 6, pp. e563–e57 7, Jun. 2021, doi: 10.1097/CCM.0000000000004916.
[29] A. E. W. Johnson *et al.*, "MIMIC-III, a freely accessible critical care database," *Sci. Data*, vol. 3, no. 1, p. 160035, May 2016, doi: 10.1038/sdata.2016.35.
[30]S. V. Buuren and K. Groothuis-Oudshoorn, "**mice**: Multivariate Imputation by Chained Equations in *R*," *J. Stat. Softw.*, vol. 45, no. 3, 2011, doi: 10.18637/jss.v045.i03.
[31] B. Bardak and M. Tan, "Using Clinical Drug Representations for Improving Mortality and Length of Stay Predictions." arXiv, Oct. 17, 2021. Accessed: Nov. 14, 2023. [Online]. Available: http://arxiv.org/abs/2110. 08918
[32] D. Weininger, "SMILES, a chemical language and information system. 1. Introduction to methodology and encoding rules," *J. Chem. Inf. Comput. Sci.*, vol. 28, no. 1, pp. 31–36, Feb. 1988, doi: 10.1021/ci00057a 005.
[33] E. E. Bolton, Y. Wang, P. A. Thiessen, and S. H. Bryant, "PubChem: Integrated Platform of Small Molec ules and Biological Activities," in *Annual Reports in Computational Chemistry*, vol. 4, Elsevier, 2008, pp. 217–241. doi: 10.1016/S1574-1400(08)00012-1.
[34]D. Duvenaud *et al.*, "Convolutional Networks on Graphs for Learning Molecular Fingerprints." arXiv, Nov. 03, 2015. Accessed: Nov. 16, 2023. [Online]. Available: http://arxiv.org/abs/1509.09292
[35] L. Breiman, "Random Forests," *Mach. Learn.*, vol. 45, no. 1, pp. 5–32, 2001, doi: 10.1023/A:1010933404 324.
[36] T. Chen and C. Guestrin, "XGBoost: A Scalable Tree Boosting System," in *Proceedings of the 22nd AC M SIGKDD International Conference on Knowledge Discovery and Data Mining*, San Francisco California USA: ACM, Aug. 2016, pp. 785–794. doi: 10.1145/2939672.2939785.
[37] G. Ke *et al.*, "LightGBM: a highly efficient gradient boosting decision tree," *Adv. Neural Inf. Process. Sy st.*, vol. 30, pp. 3146–3154, 2017.
[38] L. Prokhorenkova, G. Gusev, A. Vorobev, A. V. Dorogush, and A. Gulin, "CatBoost: unbiased boosting w ith categorical features," 2017, doi: 10.48550/ARXIV.1706.09516.